\documentclass[11pt,conference,onecolumn]{IEEEtran}
\IEEEoverridecommandlockouts
\usepackage{amsmath,amssymb,amsfonts}
\usepackage{bbm}
\usepackage[colorlinks=true, linkcolor=blue, citecolor=blue, urlcolor=blue]{hyperref}
\usepackage{cleveref}
\usepackage{caption}      
\usepackage[export]{adjustbox}  
\usepackage[most]{tcolorbox}
\usepackage{algcompatible}
\usepackage{algorithm}
\usepackage{algpseudocode}
\usepackage{graphicx}
\usepackage{varwidth} 
\usepackage{textcomp}
\usepackage{xcolor}
\usepackage{mdframed}
\usepackage{placeins}
\def\BibTeX{{\rm B\kern-.05em{\sc i\kern-.025em b}\kern-.08em
    T\kern-.1667em\lower.7ex\hbox{E}\kern-.125emX}}
\usepackage[numbers,sort&compress]{natbib}
\usepackage{comment}
\usepackage{bbm}

\begin{document}

\title{STAGED: A Multi-Agent Neural Network for Learning Cellular Interaction Dynamics}


\author{
  \IEEEauthorblockN{
    João F. Rocha\IEEEauthorrefmark{1}\IEEEauthorrefmark{4}, 
    Ke Xu\IEEEauthorrefmark{1}\IEEEauthorrefmark{4}, 
    Xingzhi Sun\IEEEauthorrefmark{1}\IEEEauthorrefmark{4}, 
    Ananya Krishna\IEEEauthorrefmark{1}, 
    Dhananjay Bhaskar\IEEEauthorrefmark{1}, \\ 
    Blanche Mongeon\IEEEauthorrefmark{2}\IEEEauthorrefmark{3}, 
    Morgan Craig\IEEEauthorrefmark{2}\IEEEauthorrefmark{3}, 
    Mark Gerstein\IEEEauthorrefmark{1},
    Smita Krishnaswamy\IEEEauthorrefmark{1}
  }
  \\
  \IEEEauthorblockA{\IEEEauthorrefmark{1}Yale University}

    \IEEEauthorblockA{\IEEEauthorrefmark{2}Sainte-Justine University Hospital Azrieli Research Centre} 
    \IEEEauthorblockA{\IEEEauthorrefmark{3}Département de mathématiques et de statistique, Université de Montréal} 
    
    \IEEEauthorblockA{\IEEEauthorrefmark{4}Equal Contribution}
}

\maketitle

\begin{abstract}

The advent of single-cell technology has significantly improved our understanding of cellular states and subpopulations in various tissues under normal and diseased conditions by employing data-driven approaches such as clustering and trajectory inference. However, these methods consider cells as independent data points of population distributions. With spatial transcriptomics, we can represent cellular organization, along with dynamic cell-cell interactions that lead to changes in cell state. Still, key computational advances are necessary to enable the data-driven learning of such complex interactive cellular dynamics. While agent-based modeling (ABM) provides a powerful framework, traditional approaches rely on handcrafted rules derived from domain knowledge rather than data-driven approaches. To address this, we introduce \textbf{S}patio \textbf{T}emporal \textbf{A}gent-Based \textbf{G}raph \textbf{E}volution \textbf{D}ynamics (STAGED) integrating ABM with deep learning to model intercellular communication, and its effect on the intracellular gene regulatory network. Using graph ODE networks (GDEs) with shared weights per cell type, our approach represents genes as vertices and interactions as directed edges, dynamically learning their strengths through a designed attention mechanism. Trained to match continuous trajectories of simulated as well as inferred trajectories from spatial transcriptomics data, the model captures both intercellular and intracellular interactions, enabling a more adaptive and accurate representation of cellular dynamics.

\end{abstract}

\section{Introduction}

Single-cell technology enabled researchers to understand different cellular states and subpopulations under different conditions through clustering, denoising and visualization~\citep{Louvain_Blondel2008,LeidenWaltman2013,t-SNE_vandermaaten08a,UMAPMcInnes2018,PHATEMoon2019}. Moreover, modern neural optimal transport methods like MIOFlow~\cite{huguet2022manifold}, OT-CFM~\cite{OT-CFM} and Trajectorynet~\cite{TrajectoryNET} are capable of deriving single cell trajectories from static snapshot data. However, these methods treat cells as entities that exist in isolation, as if their state evolve independently, disregarding critical cellular interactions that drive biological dynamics - a limitation inherent to single-cell data, which provides high-dimensional readouts but lacks spatial context.

\begin{figure}[htbp]
\centering
\includegraphics[width=1.0\linewidth]{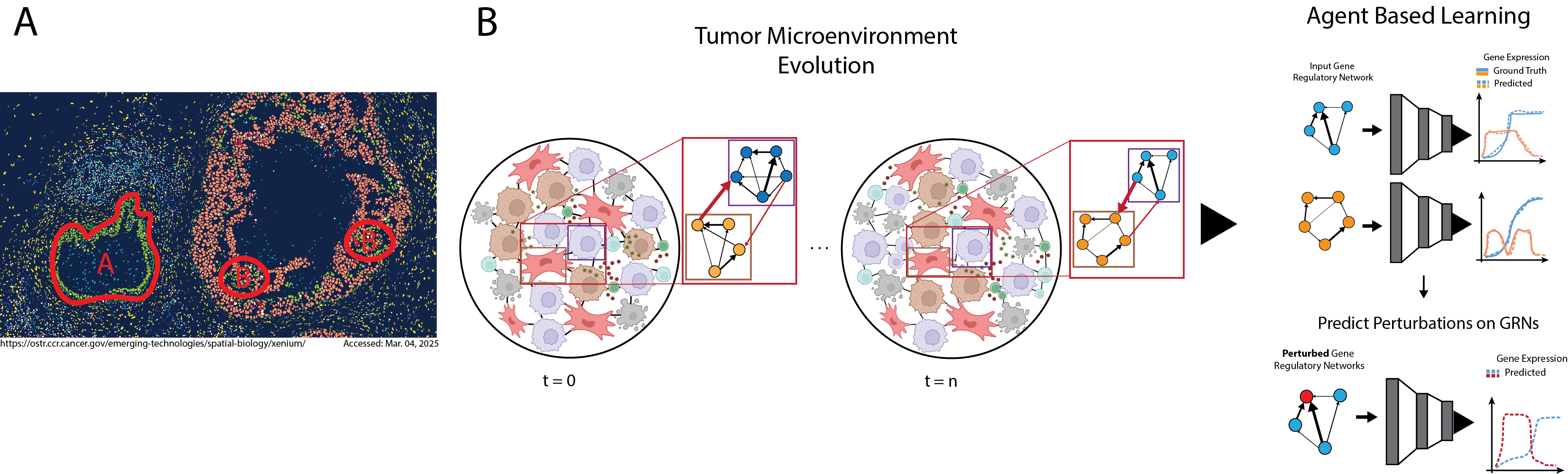}
\caption{
\textbf{STAGED infers regulatory and cell–cell interaction networks using spatial transcriptomics data.} \textbf{A.} Spatial transcriptomics data from Xenium \citep{10xGenomics_Xenium_2025}. Cells in regions A and B express similar genes but occupy distinct spatial neighborhoods. Traditional single-cell methods, which lack spatial context, treat such cells as equivalent, ignoring important differences in their local signaling landscapes. \textbf{B.} STAGED models each cell as an agent governed by a graph neural ODE, capturing dynamic gene regulation and communication with nearby cells through ligand–receptor signaling. STAGED uses graph attention to refine both gene–gene and cell–cell interactions over time and is trained to predict gene expression dynamics based on these evolving networks. It supports downstream analysis such as predicting the effects of perturbations to gene regulatory networks.}
\label{fig:stranscriptomics}
\end{figure}

Spatial transcriptomics methods \citep{chen2015spatially,janesick2023high,rao2020bridging,moses2022museum}, address this limitation by preserving the spatial organization of cells within a tissue (Figure \ref{fig:stranscriptomics}A). This information is crucial since cellular behavior, development and function are inherently dependent on inter-cellular communication and the local microenvironment. However, existing computational approaches fail to fully leverage spatial transcriptomics, as they do not adequately model cells as dynamic entities within a complex system of interactions \citep{rao2021exploring,williams2022introduction,tian2023expanding,sun2024hyperedge}. Agent-Based Modeling (ABM), provides a powerful framework for modeling cells as interacting agents in space ~\citep{Monti2023,mongeon2024spatial} but they rely on hand-crafted rules derived from domain knowledge, which might introduce bias compared to a data-driven approach.

To address these limitations, we developed STAGED (\textbf{S}patio- \textbf{T}emporal \textbf{A}gent-Based \textbf{G}raph \textbf{E}volution \textbf{D}ynamics), a deep neural network composed of multiple interacting agents that learns both intra-cellular regulatory networks and inter-cellular interactions directly from spatial transcriptomics data. STAGED operates on two interconnected levels: (1) intercellular communication, where agents exchange signals with spatial neighbors via ligand–receptor interactions, and (2) intracellular regulation, where gene–gene interactions within each agent are inferred and updated continuously over time (Figure \ref{fig:stranscriptomics}B). Each agent in STAGED corresponds to a single cell whose internal gene expression dynamics are governed by a graph neural ODE (GDE)~\citep{chen2018neural,poli2019graph,jin2022multivariate,bhaskar2024inferring}. Genes are represented as nodes in a directed graph, with regulatory influences encoded as time-varying edges. These edge weights evolve according to a graph attention mechanism~\citep{vaswani2017attention,velivckovic2017graph} that incorporates historical gene expression to capture time-delayed transcriptional and signaling effects. Agents also exchange signals with spatial neighbors through ligand–receptor interactions. To reflect conserved regulatory logic, agents of the same cell type share GDE parameters while still exhibiting context-dependent behavior based on their spatial neighborhood and neighboring cell states.

We evaluate STAGED across two simulation settings and one real-world dataset. First, STAGED captures population-level dynamics and cell-cell interactions in simulations of glioblastoma where an oncolytic virus induces interactions among heterogeneous cell populations. Second, we simulate oscillatory gene expression dynamics and demonstrate STAGED's ability to accurately recover the underlying time-varying regulatory interactions. Finally, we apply STAGED to spatial transcriptomics data capturing early Alzheimer’s disease progression, using data from the Seattle Alzheimer’s Disease Brain Cell Atlas (SEA-AD)~\cite{hawrylycz_sea-ad_2024}. Here, STAGED recovers spatially localized and cell-type-specific gene regulatory networks, demonstrating its utility in real-world disease contexts. To summarize, our \emph{main contributions} are:

\begin{itemize}
\item We developed STAGED, a deep neural network composed of multiple agents, each implementing a graph neural ODE, for inferring intra- and inter-cellular dynamics from spatial transcriptomics data.
\item STAGED incorporates a two-level attention mechanism: (i) at the tissue level, to model dynamic ligand–receptor-mediated signaling between neighboring cells, and (ii) at the subcellular level, to learn time-varying gene–gene regulatory interactions.
\item STAGED uses a history-aware attention mechanism within its GDEs for both gene–gene and cell–cell interactions. By incorporating past and present values when computing attention weights, the model captures realistic time delays in transcription, intracellular signaling, and cell–cell communication.
\item To evaluate STAGED, we developed a simulation environment that generates realistic spatially resolved gene expression trajectories with controllable intracellular regulatory logic and cell–cell signaling.
\end{itemize}

\vspace{-10pt}

\section{Background and Related Work}

Here we review two key frameworks that serve as precursors to STAGED:  MIOFlow~\citep{huguet2022manifold} and RITINI~\citep{bhaskar2024inferring}. These methods respectively address trajectory inference and regulatory network learning from single-cell data, but do not incorporate intercellular interactions. RITINI pioneered the idea of using attention-based GDEs to learn gene regulatory interactions, using MIOFlow derived gene expression trajectories as training data. STAGED builds upon this, incorporating GDEs within an agent-based framework to simultaneously infer regulatory interactions and spatiotemporal cellular communication while learning to predict gene expression trajectories.

MIOFlow~\citep{huguet2022manifold} (Manifold Interpolating Optimal-Transport Flows) infers continuous single-cell trajectories from static snapshot data by modeling the dynamics with a neural ODE network~\citep{chen2018neural,liu2024imageflownet,sun2024deep,borsari2023chronode} that outputs time-varying derivatives of the trajectories $f_\theta (x_t,t)$, which is then integrated to create the transport path (or trajectory) of the cell. MIOFlow transports populations of cells over time using dynamic optimal transport~\citep{villani2008optimal,santambrogio2015optimal,tong2020trajectorynet}, such that the paths between cells collected at different timepoints (or cells binned into different pseudotimes) are as efficient and smooth as possible. The end result of MIOFlow is cell trajectories through the transcriptomic state space, showing how individual cells evolve by changing their transcriptomic profiles. For spatiotemporal data, MIOFlow can be naturally extended to also predict the spatial trajectories of cells over time.

RITINI~\citep{bhaskar2024inferring} (Regulatory Temporal Interaction inference) complements MIOFlow by directly modeling the regulatory graph within the GDE. Starting with a prior graph based on known gene-gene interaction relationships as an initial graph, RITINI refines the network with a space-and-time attention mechanism. The attention-weights of the dynamic graph between genes vary with time, allowing new connections to emerge. RITINI is trained to match the gene expression inferred by MIOFlow models the dynamics of the features.

Thus, MIOFlow and RITINI work together, with the former network learning a trajectory through the transcriptional state space, and the latter learning the regulatory networks that can produce the corresponding transcriptional dynamics. 

However, this paradigm does not model cell-cell interactions, crucial for understanding the biological system. We note that including cell-cell interaction into this style of model requires a significant paradigm change to create an agent-based system with several cellular GDE agents that interact with one another over time. 

\section{Methods}

\begin{figure}[htbp]
    \centering
\includegraphics[width=\linewidth]{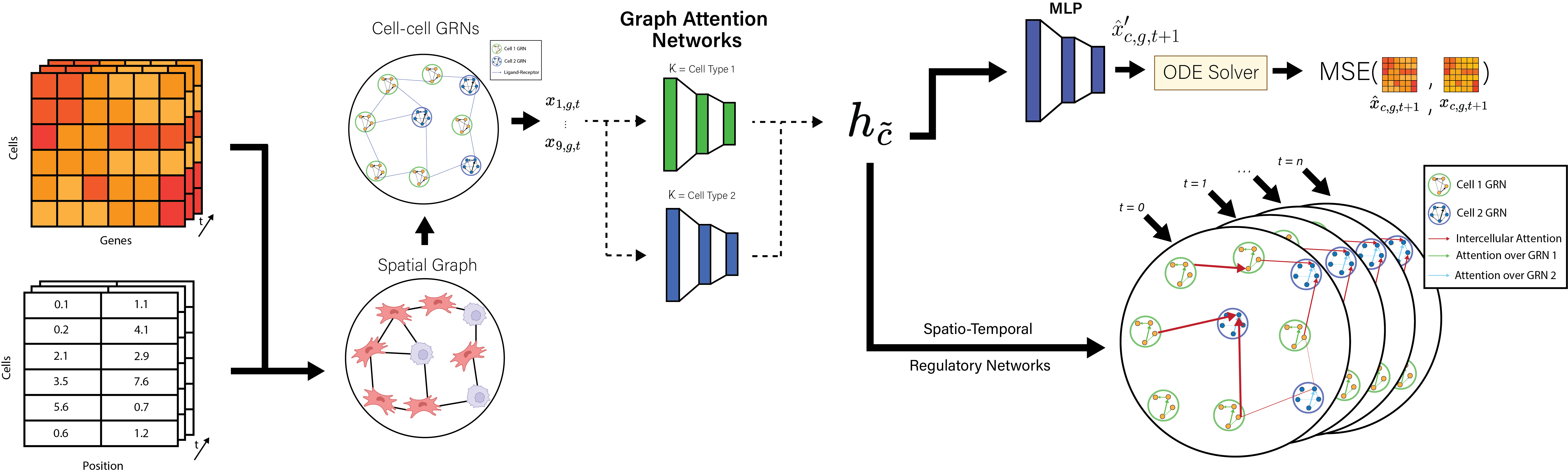}
    \caption{\textbf{STAGED model}: Each agent in STAGED is a cell-type–specific model composed of a graph attention network (GAT) and a neural ODE. The GAT aggregates information from spatially proximal cells via ligand–receptor–mediated interactions, producing a context-aware embedding of each gene. This embedding is passed through a multi-layer perceptron (MLP) to predict the gene’s time derivative, which the ODE solver then integrates to forecast future gene expression levels. Ultimately, our objective is to extract Spatio-Temporal Regulatory Networks from the learned attention weights, enabling novel biological insights.}
    \label{fig:model}
\end{figure}

We present STAGED, a multi-agent graph-based framework for modeling the spatiotemporal dynamics of both cell–cell interactions and intracellular gene regulatory networks (GRNs) within a tissue. STAGED operates on two interdependent dynamic graphs: a tissue-level graph, where each vertex represents a cell and edges reflect spatial proximity-based neighborhoods that evolve over time; and a cell-level graph embedded within each cell, where nodes correspond to genes and edges denote gene regulatory interactions, also dynamically updated (see Figure \ref{fig:model}).

Both graphs are time-varying, enabling the model to capture spatial reorganization of cells and context-dependent gene regulation. The tissue-level graph is constructed on-the-fly at each time step based on cell positions.

Each cell in the tissue-level graph hosts an agent - a Graph Neural ODE (GDE) - that models the dynamics of its internal GRN. These agents interact via ligand-receptor mediated message passing across neighboring cells, allowing the model to capture how intercellular communication modulates intracellular gene expression. By predicting temporal changes in gene expression across all cells, STAGED uncovers regulatory and signaling mechanisms driving spatially coordinated changes in cell fate.

\subsection{Predicting Multicellular Gene Expression Dynamics via Regulatory Network Graphs}

We now formalize the problem setup. Let
$\mathcal{D} = \{(x_{c,g,t}, y_{c,t}) : c \in C,\, g \in G,\, t = 0, \dots, T\}$
denote the spatiotemporal dataset, where $C$ is the set of cells, $G$ is the set of genes, and $t$ indexes discrete time points. Here, $x_{c,g,t}$ represents the expression of gene $g$ in cell $c$ at time $t$, and $y_{c,t}$ denotes the spatial location of cell $c$ at time $t$. We define the gene expression and spatial location matrices at time $t$ as:
$\mathbf{x}_t = (x_{c,g,t})_{c \in C,\, g \in G}, \mathbf{y}_t = (y_{c,t})_{c \in C}.$

%

Our main goal in this work is to infer a time-varying interaction graph via the proxy task of matching time trajectories of cellular gene expression (Figure \ref{fig:model}). This is achieved by designing a model $m$ that, given the initial gene expression $\mathbf{x}_0$ and the sequence of spatial locations $\mathbf{y}_{0:T} = (\mathbf{y}_0, \dots, \mathbf{y}_T)$, predicts the future gene expression values:
$
\hat{\mathbf{x}}_{1:T} = (\hat{\mathbf{x}}_1, \dots, \hat{\mathbf{x}}_T) = m(\mathbf{x}_0, \mathbf{y}_{0:T}).
$
To model dynamic dependencies among genes, receptors, and ligands across time and neighboring cells, we use a graph attention mechanism within $m$. At each time step, an interaction graph is built from spatial context and biological priors. A single-head graph attention layer updates node features by weighting edges based on learned interaction relevance. The model is trained by minimizing the mean squared error loss between $\mathbf x_{1:T}$ and $\hat{\mathbf x}_{1:T}$.

\subsection{Architecture of STAGED}
Let $\mathbf{x}_{c,t} = (x_{c,g,t})_{g \in G}$ denote the gene expression vector for cell $c$ at time $t$.  We define a set of GAT modules $\{h_{\phi_k} : k \in \mathcal{K}\}$, one per cell type, where $\phi_k$ are the parameters associated with cell type $k$.  Each module produces an embedding $h_{\phi_k}(\mathbf{x}_{c,t}, t)$ for cell $c$ based on its expression profile and time, with $k$ denoting the type of cell $c$ (i.e., $c \in C_k$).

We extend these to \textit{STAGED modules} $\{a_{\theta_c,\phi_c}(\mathbf{x}_t, \mathbf{y}_t, t) : c \in C\}$ by incorporating information from neighboring cells via an attention mechanism. Specifically, the attention coefficient $\beta_{ij}^{(cs)}$ quantifies the influence from gene $j$ in cell $s$ to gene $i$ in cell $c$, as computed by a graph attention network. The aggregated information is concatenated and fed into a multi-layer perceptron $f_{\theta_c}$ (parameterized by $\theta_c$) that predicts the derivative for the GDE at time $t$. Formally, we define:
$
a_{\theta_c,\phi_c}(\mathbf{x}_t, \mathbf{y}_t, t)
= f_{\theta_c} \Biggl(  \, h_{\phi_c}(\mathbf{x}_{c,t}, t) \Vert \sum_{s \in N(c \mid \mathbf{y}_t)} \sum_{i,j \in G} \beta_{ij}^{(cs)}\, x_{c,i,t}\, x_{s,j,t-\delta} \Vert t \Biggr),
$
where $\Vert$ denotes concatenation, $\delta$ is the signaling time lag, and the dynamic neighborhood of cell $c$ is defined as
$
N(c \mid \mathbf{y}_t) = \{ s \in C : \|y_{s,t} - y_{c,t}\| \leq r \}.
$
This process constructs a dynamic graph $\mathcal G(t)$ by connecting cells within a radius $r$ 
to capture local cell-cell interactions.
Our framework is illustrated in Figure \ref{fig:stranscriptomics}B.

\vspace{2pt}
\noindent For full algorithmic details, including node feature assignments and temporal lags, see Algorithm~\ref{alg:algorithm} in the Appendix.

\subsection{Biological Priors via Weight-Sharing \& Masked Attention}
We incorporate two key biological priors into our model:

\textbf{(i) Weight-Sharing:}  
Cells of the same type (e.g., immune cells or tumor cells) typically perform similar functions. To reflect this, we partition the cell set into types:
$
C = C_1 \sqcup C_2 \sqcup \dots \sqcup C_K.
$
For every cell $c \in C_i$, we enforce weight-sharing by setting
$
\phi_c \equiv \phi_i \; \text{and} \; \theta_c \equiv \theta_i.
$
This reduces the number of distinct STAGED module parameters to $K$. The combination of the attention and weight sharing enforces similarity between cells of the same type, while allowing context and neighborhood dependent changes in behavior. Thus the same cell type can have different trajectories based on the cellular neighborhood. 

\textbf{(ii) Masked Attention:}  
Since intercellular signals are transmitted via ligand--receptor pairs, we constrain the attention mechanism to focus only on relevant genes. Specifically, we apply an attention mask such that
$
\beta_{ij}^{(cs)} \leftarrow \beta_{ij}^{(cs)} \, \mathbbm{1}(j \in L \text{ and } i \in R),
$
where $L, R \subset G$ denote the sets of ligand and receptor genes, respectively.

\section{Results}

To evaluate STAGED we used two biologically grounded simulation environments and a real-world dataset to assess its ability to learn dynamic, spatially-resolved gene regulatory programs. 

\subsection{STAGED learns multicellular interactions in a tumor microenvironment simulation}

\begin{figure}[htbp]
\centering
\includegraphics[width=\linewidth]{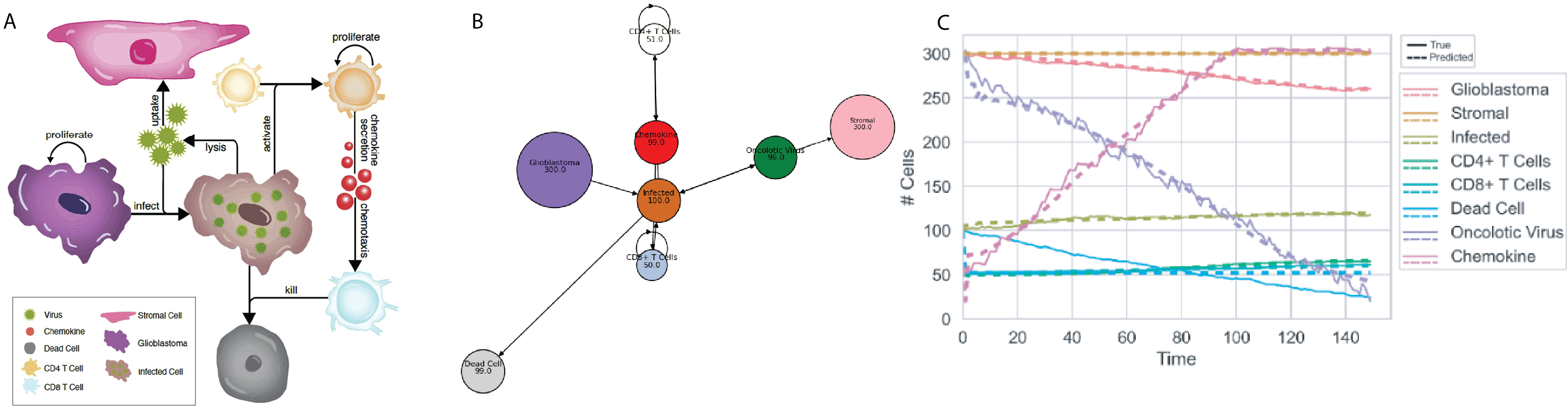}
\caption{\textbf{Agent-based simulation of glioblastoma} \textbf{A.} Agent-based simulation model for a glioblastoma in an oncolitic virus environment from \cite{mongeon2024spatial}. \textbf{B.} Graph is used to maintain information on the overall state of the agent. \textbf{C.} STAGED was able to fully recover population dynamics.}
\label{fig:glioblastoma}
\end{figure}

First, we validated with a model of Glioblastoma Tumour Microenvironment agent-based simulation proposed by \cite{mongeon2024spatial} (Figure \ref{fig:glioblastoma}). We defined the cell as agents and their different interactions as described in Figure \ref{fig:glioblastoma}A. Agents were modeled as nodes in a graph and could perform different actions through the edges (Figure \ref{fig:glioblastoma}B). Here, the state is not yet the GRN of each agent, but a simplified version of proliferation rates, death, and total count features. The model retrieved the total count of agents in the environment and captured the varying dynamics of each agent (Figure\ref{fig:glioblastoma}C).

\subsection{STAGED infers regulatory networks that exhibit complex oscillatory dynamics}

In the second setting, we tested STAGED's ability to infer complex gene regulatory dynamics by simulating intercellular oscillatory gene expression patterns within a spatially distributed population of cells. 

We designed a new environment using the Gillespy2~\cite{matthew2023gillespy2} biochemical systems simulator for this purpose. We defined two different cell types (cancer and stromal). Each cell type has an initial default and a position. From this GRN, we connect the cells to their k-nearest neighbors in Euclidean space. Then we use a list of ligand-receptors to connect the cell genes to its neighbors' genes inside the simulator. This creates a model that contains all cell genes as entities and their interactions defined by stochastic functions. The functions could be activation or repression. 

To simulate gene expression dynamics within a regulatory network, we employ a framework based on ordinary differential equations (ODEs). Each gene is modeled as a continuous-valued species in GillesPy2, with its expression level evolving over time in response to regulatory influences from both activators and repressors. Activating effects are represented using Hill functions to capture nonlinear, saturating behavior, while repression is modeled linearly. The net regulatory input determines the direction of regulation - either upregulation or downregulation - as described in equation~\eqref{eq:gene_dynamics}. The system dynamics are governed by a continuous ODE formulation~\eqref{eq:gene_dynamics_ODE} that smoothly approximates piecewise regulatory behavior to ensure biologically plausible bounded expression levels. Upregulation occurs when the net regulatory input is positive, downregulation when negative, and a baseline decay term ensures stability in the absence of strong regulatory signals. Expression levels are constrained by a gene-specific maximum \( x_g^{\max} \).

The regulatory input \( R_g(x) \) for gene \( g \) aggregates contributions from its activators \( \mathcal{A}_g \) and repressors \( \mathcal{R}_g \). For each activator \( a \in \mathcal{A}_g \), \( x_a \) denotes its expression level, \( k_{a \to g} \) the activation strength (\texttt{rate\_constant}), \( \theta_{a \to g} \) the activation threshold (default 0.5), and \( n_{a \to g} \) the Hill coefficient (default 2). For each repressor \( r \in \mathcal{R}_g \), \( x_r \) is its expression level, and \( k_{r \to g} \) its repression strength (\texttt{rate\_constant}). The function \( \mathrm{ReLU}(z) = \max(0, z) = \frac{z + |z|}{2} \) is used to differentiate between positive and negative contributions to gene regulation. The parameter \( \delta_g \) denotes the passive decay rate for gene \( g \) (\texttt{deactivation\_rate}), governing expression loss when regulatory input is weak or near zero.

\begin{figure}[]
\centering
\includegraphics[width=\linewidth]{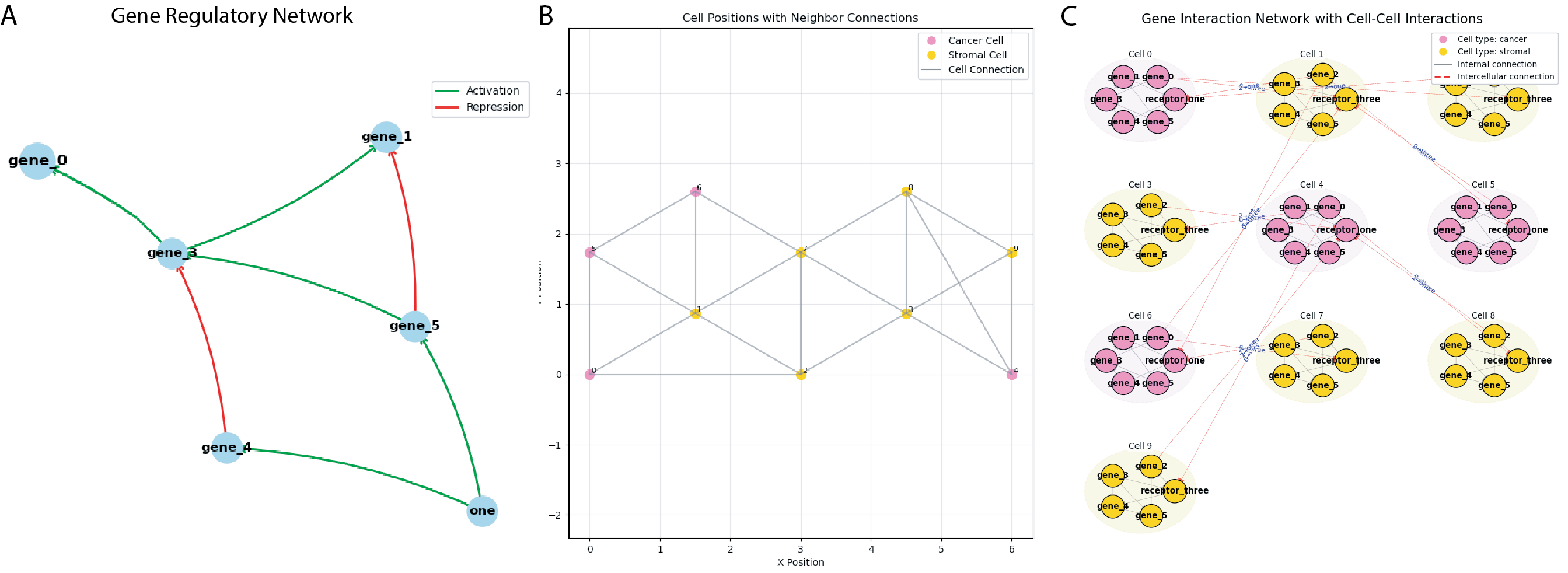}
\caption{\textbf{Spatially resolved stochastic simulation of gene expression dynamics in a randomly constructed gene regulatory network (GRN) using GillesPy2.} \textbf{A.} Ground truth GRN where nodes represent genes, directed red edges indicate repression and directed green edges indicate activation. \textbf{B.} Tissue-level graph showing the positioning of simulated cells in space. \textbf{C.} Ground truth network consisting of cell-cell communication between neighboring cells mediated by ligand-receptor interactions, with the intrinsic dynamics of each cell controlled by the GRN.}
\label{fig:Simulation}
\end{figure}

We illustrate the construction of the simulated system used to train and evaluate STAGED in three layers, each capturing increasing biological complexity (Figure~\ref{fig:Simulation}).

First, we define the gene regulatory networks (GRNs) for each cell type (Figure~\ref{fig:Simulation}A), where nodes represent genes and edges denote regulatory relationships—either activation or repression—parameterized. These GRNs govern the internal expression dynamics within each cell. For each GRN, we simulate the temporal evolution of gene expression using the ODEs, generating characteristic trajectories such as oscillations, saturations, or decay depending on the network topology and regulatory logic. Second (Figure~\ref{fig:Simulation}B), we place cells in a two-dimensional space and establish neighborhood connections based on spatial proximity. Each cell is assigned a type, and edges between nearby cells define potential communication pathways, capturing the influence of spatial organization. Finally (Figure~\ref{fig:Simulation}C), we incorporate intercellular signaling into the GRNs using predefined ligand-receptor pairs. This step extends the cell-intrinsic networks to include regulatory edges across neighboring cells, enabling the simulation of coupled gene expression dynamics influenced by both cell-intrinsic regulation and cell-cell communication.

This modeling framework captures key biological features, including saturating activation, linear repression, and passive decay, while maintaining interpretability and bounded expression dynamics. The full network is compiled into a GillesPy2 model and simulated using deterministic ODE solvers, allowing flexible specification of gene-specific parameters, network topology, and regulatory logic. This enables mechanistically grounded and interpretable simulations of gene expression over time.

\begin{equation}
R_g(x) = \sum_{a \in \mathcal{A}_g} \frac{k_{a \to g} \cdot x_a^{n_{a \to g}}}{\theta_{a \to g}^{n_{a \to g}} + x_a^{n_{a \to g}}}
- \sum_{r \in \mathcal{R}_g} k_{r \to g} \cdot x_r
\label{eq:gene_dynamics}
\end{equation}

\begin{equation}
\frac{dx_g}{dt} =
\mathrm{ReLU}(R_g(x)) \cdot \left(1 - \frac{x_g}{x_g^{\max}}\right)
- \mathrm{ReLU}(-R_g(x)) \cdot x_g
+ \mathrm{ReLU}(1 - |R_g(x)|) \cdot \delta_g \cdot x_g
\label{eq:gene_dynamics_ODE}
\end{equation}

\begin{figure}[]
    \centering
    \includegraphics[width=\linewidth]{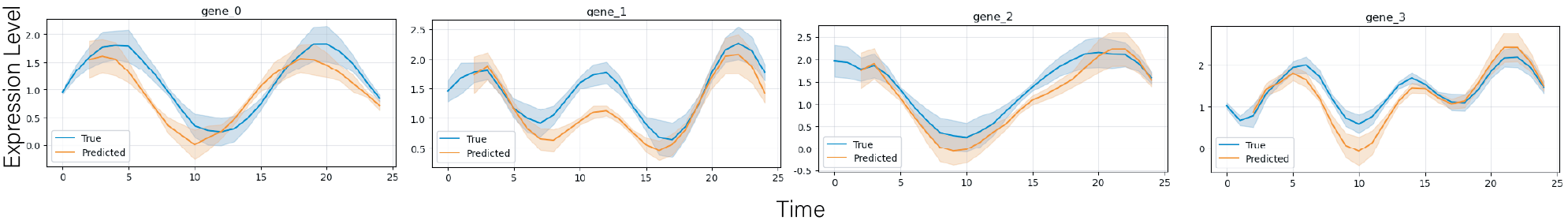}
    \caption{\textbf{STAGED accurately predicts complex gene expression dynamics.} We simulated a population of 7 cells of 2 different cell types, each consisting of a small regulatory network of 6 genes that exhibited oscillatory dynamics. Cells were coupled by spatial proximity and Gaussian noise was added to the simulated gene expression trajectories. Cells exhibited out-of-phase oscillatory dynamics. STAGED was able to accurately capture these dynamics, shown here one cell type for a subset of genes. Mean and standard deviation are calculated using all cells of the type.
    }
    \label{fig:gene_expression_time}
\end{figure}

To test STAGED in a more complex setting, we simulated oscillatory dynamic microenvironments of two cell types and seven cells. Specifically, we constructed two different types of GRNs, each containing six genes. STAGED successfully reconstructed the oscillatory dynamics for each agent (Figure \ref{fig:gene_expression_time}), confirming the model's ability to learn expressive representations of internal regulatory networks. 

Beyond quantitative prediction accuracy, STAGED provides interpretable insights into both intra- and inter-cellular interactions. The learned attention maps reveal context-specific regulatory graphs and gene expression for each cell type, uncovering key signaling interactions (e.g., ligand-receptor pairs) and gene-gene dependencies modulated by the spatial information (Figure \ref{fig:infered_grn}). These attention-derived graphs not only reflect biological priors encoded during training, but also adapt to dynamically changing spatial contexts, suggesting STAGED’s utility as a hypothesis-generating tool for discovering novel regulatory mechanisms.

\begin{figure}[]
\centering
\includegraphics[width=\linewidth]{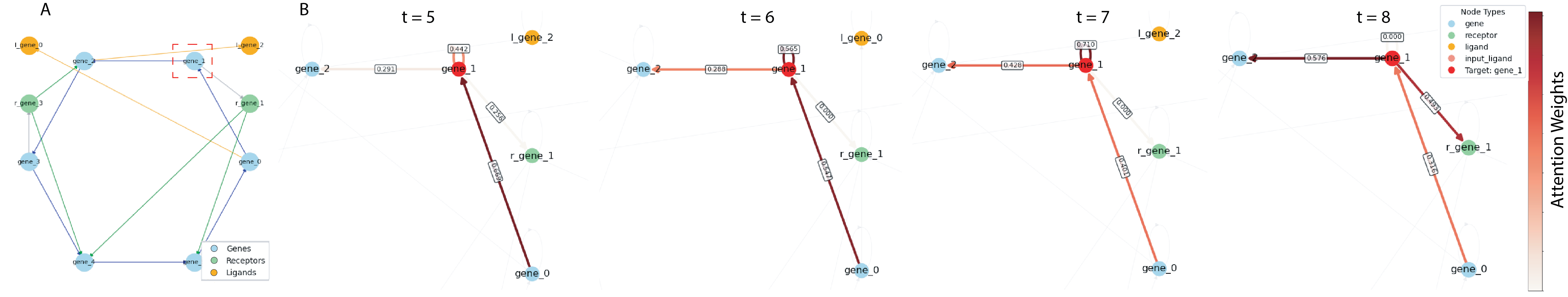}
\caption{\textbf{Temporally varying regulatory interactions inferred using the attention mechanism in STAGED.} \textbf{A.} Simulated gene regulatory network (GRN) exhibiting oscillatory dynamics, where genes activate sequentially in a circular pattern. \textbf{B.} Time-resolved gene-gene interaction profiles inferred for gene 1 using attention weights from STAGED. The inferred interactions recapitulate the simulated cyclical activation pattern, demonstrating STAGED's ability to capture temporally varying regulatory relationships.}
\label{fig:infered_grn}
\end{figure}

\subsection{STAGED uncovers distinct regulatory programs in microglia during Alzheimer’s disease}

Having validated STAGED on simulated datasets, we next applied it to spatial transcriptomics data from the Seattle Alzheimer’s Disease Brain Cell Atlas (SEA-AD) \citep{hawrylycz_sea-ad_2024}, a multimodal resource that profiles the aging human brain with an emphasis on early-stage Alzheimer’s disease.

We focused on microglia, a glial cell type implicated in the early progression of Alzheimer’s. Using spatial coordinates and cell-type annotations provided in SEA-AD, we extracted microglial cells and performed unsupervised clustering, identifying eight transcriptionally distinct subpopulations.

STAGED was then trained on the microglial cell population to learn gene expression dynamics conditioned on spatial context. To evaluate model performance, we compared predicted and observed expression trajectories for representative genes in each cluster using Spearman correlation. STAGED accurately captured gene expression profiles across clusters, with strong agreement between predicted and observed trajectories for representative genes (Figure~\ref{fig:infered_grn}A).

Importantly, analysis of the learned attention weights revealed striking heterogeneity in inferred gene-gene interactions across microglial subpopulations (Figure~\ref{fig:infered_grn}B). Each cluster was associated with a distinct regulatory program, centered around different key genes. This suggests that microglia may occupy diverse regulatory states within the same tissue - an insight that could be leveraged to design more precise, cell-state-specific therapeutic strategies. 

\begin{figure}[]
    \centering
    \includegraphics[width=0.88\linewidth]{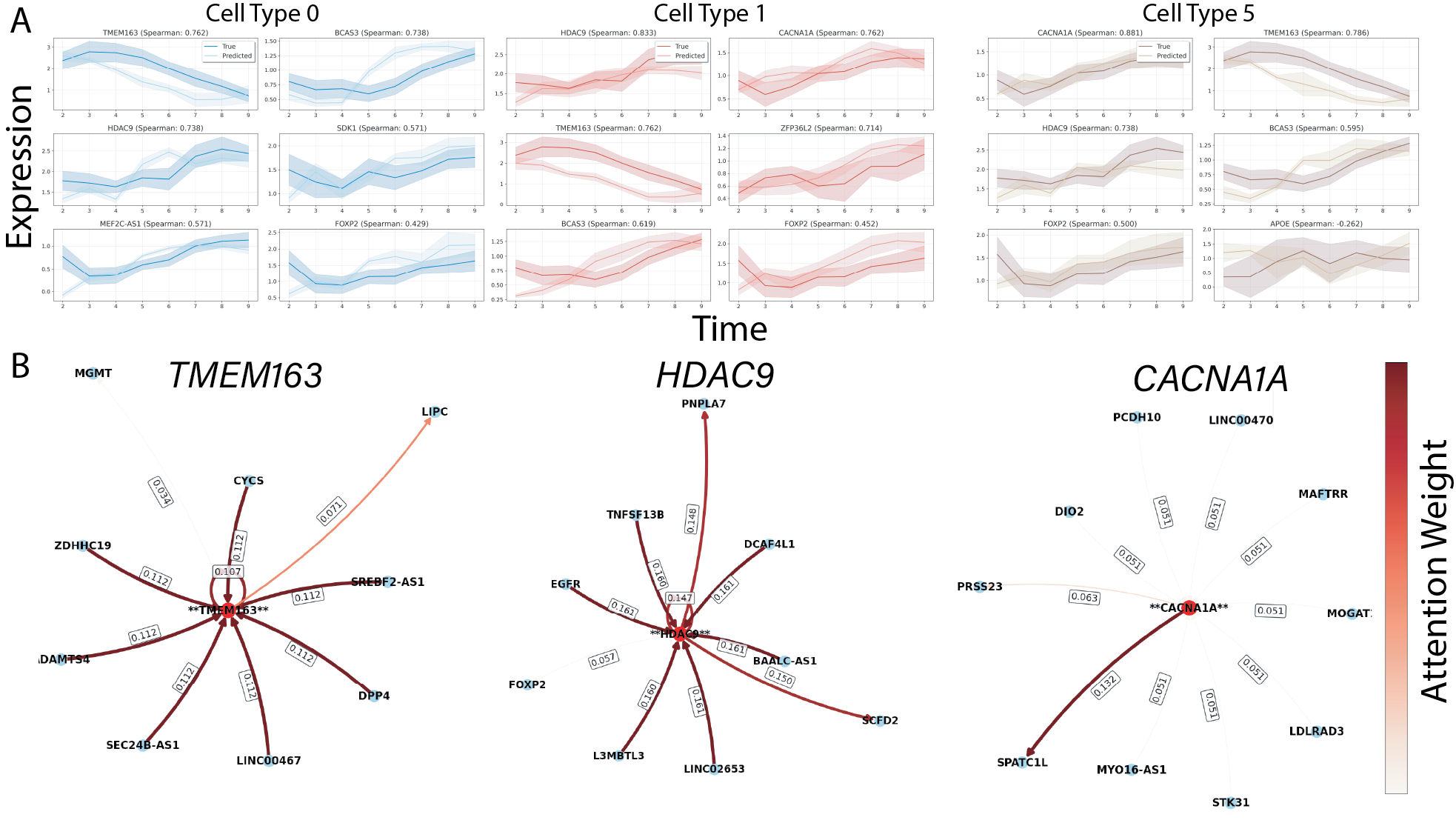}
    \caption{\textbf{STAGED identifies cell-type-specific gene dynamics and regulatory networks in microglia subpopulations. }\textbf{A.} Top expressed genes across three distinct microglia subpopulations. STAGED recovers the gene dynamics and focus in different genes for each of the cell types. \textbf{B.} For each microglial subpopulation, the network is centered on the top-ranking gene identified in panel A, showing how these key genes interact.}
    \label{fig:infered_grn}
\end{figure}

\section{Conclusion}

We developed STAGED, a novel framework that infers both intra- and inter-cellular regulatory networks from spatial transcriptomics data. Unlike prior approaches that focus exclusively on intracellular gene regulatory networks, STAGED leverages attention-driven graph neural ODEs to jointly learn transcriptional dynamics and cell-cell communication via ligand–receptor signaling.

A key innovation of STAGED lies in its time-varying attention mechanisms at both the cellular and gene levels. These attention maps evolve dynamically over time, enabling the model to uncover how transcriptional regulation and intercellular signaling are modulated by spatial context and over time. Furthermore, the incorporation of history-aware attention with explicit time lags allows STAGED to capture biologically realistic delays associated with transcriptional regulation, intracellular signaling cascades, and extracellular communication mechanisms such as ligand diffusion and receptor binding.

Through applications to simulated tumor microenvironments, synthetic circuits exhibiting oscillatory dynamics, and spatial transcriptomics data of Alzheimer's disease progression, we demonstrated that STAGED faithfully reconstructs dynamic gene expression trajectories and reveals interpretable, context-dependent regulatory programs.

STAGED bridges the gap between mechanistic simulation and machine learning by offering a scalable, interpretable, and biologically grounded platform for modeling multicellular systems. Its ability to integrate spatiotemporal dynamics, history-aware attention, and ligand–receptor signaling positions it as a powerful tool for hypothesis generation, regulatory network discovery, and \textit{in silico} experimentation. Future directions include modeling epigenetic regulation, integrating multimodal spatial omics data, and simulating perturbations such as gene knockouts, receptor inhibition, and microenvironmental remodeling to investigate causal mechanisms in development and disease.

\clearpage

\bibliography{mlcb_references}

\begin{thebibliography}{10}

\bibitem{Louvain_Blondel2008}
Vincent~D Blondel, Jean-Loup Guillaume, Renaud Lambiotte, and Etienne Lefebvre.
\newblock Fast unfolding of communities in large networks.
\newblock {\em Journal of Statistical Mechanics: Theory and Experiment}, 2008(10):P10008, October 2008.

\bibitem{LeidenWaltman2013}
Ludo Waltman and Nees~Jan van Eck.
\newblock A smart local moving algorithm for large-scale modularity-based community detection.
\newblock {\em The European Physical Journal B}, 86(11), November 2013.

\bibitem{t-SNE_vandermaaten08a}
Laurens van~der Maaten and Geoffrey Hinton.
\newblock Visualizing data using t-sne.
\newblock {\em Journal of Machine Learning Research}, 9(86):2579--2605, 2008.

\bibitem{UMAPMcInnes2018}
Leland McInnes, John Healy, Nathaniel Saul, and Lukas Großberger.
\newblock Umap: Uniform manifold approximation and projection.
\newblock {\em Journal of Open Source Software}, 3(29):861, September 2018.

\bibitem{PHATEMoon2019}
Kevin~R. Moon, David van Dijk, Zheng Wang, Scott Gigante, Daniel~B. Burkhardt, William~S. Chen, Kristina Yim, Antonia van~den Elzen, Matthew~J. Hirn, Ronald~R. Coifman, Natalia~B. Ivanova, Guy Wolf, and Smita Krishnaswamy.
\newblock Visualizing structure and transitions in high-dimensional biological data.
\newblock {\em Nature Biotechnology}, 37(12):1482–1492, December 2019.

\bibitem{huguet2022manifold}
Guillaume Huguet, Daniel~Sumner Magruder, Alexander Tong, Oluwadamilola Fasina, Manik Kuchroo, Guy Wolf, and Smita Krishnaswamy.
\newblock Manifold interpolating optimal-transport flows for trajectory inference.
\newblock {\em Advances in neural information processing systems}, 35:29705--29718, 2022.

\bibitem{OT-CFM}
Alexander Tong, Kilian Fatras, Nikolay Malkin, Guillaume Huguet, Yanlei Zhang, Jarrid Rector-Brooks, Guy Wolf, and Yoshua Bengio.
\newblock Improving and generalizing flow-based generative models with minibatch optimal transport, 2023.

\bibitem{TrajectoryNET}
Alexander Tong, Jessie Huang, Guy Wolf, David van Dijk, and Smita Krishnaswamy.
\newblock Trajectorynet: A dynamic optimal transport network for modeling cellular dynamics, 2020.

\bibitem{chen2015spatially}
Kok~Hao Chen, Alistair~N Boettiger, Jeffrey~R Moffitt, Siyuan Wang, and Xiaowei Zhuang.
\newblock Spatially resolved, highly multiplexed rna profiling in single cells.
\newblock {\em Science}, 348(6233):aaa6090, 2015.

\bibitem{janesick2023high}
Amanda Janesick, Robert Shelansky, Andrew~D Gottscho, Florian Wagner, Stephen~R Williams, Morgane Rouault, Ghezal Beliakoff, Carolyn~A Morrison, Michelli~F Oliveira, Jordan~T Sicherman, et~al.
\newblock High resolution mapping of the tumor microenvironment using integrated single-cell, spatial and in situ analysis.
\newblock {\em Nature communications}, 14(1):8353, 2023.

\bibitem{rao2020bridging}
Nikhil Rao, Sheila Clark, and Olivia Habern.
\newblock Bridging genomics and tissue pathology: 10x genomics explores new frontiers with the visium spatial gene expression solution.
\newblock {\em Genetic Engineering \& Biotechnology News}, 40(2):50--51, 2020.

\bibitem{moses2022museum}
Lambda Moses and Lior Pachter.
\newblock Museum of spatial transcriptomics.
\newblock {\em Nature methods}, 19(5):534--546, 2022.

\bibitem{rao2021exploring}
Anjali Rao, Dalia Barkley, Gustavo~S Fran{\c{c}}a, and Itai Yanai.
\newblock Exploring tissue architecture using spatial transcriptomics.
\newblock {\em Nature}, 596(7871):211--220, 2021.

\bibitem{williams2022introduction}
Cameron~G Williams, Hyun~Jae Lee, Takahiro Asatsuma, Roser Vento-Tormo, and Ashraful Haque.
\newblock An introduction to spatial transcriptomics for biomedical research.
\newblock {\em Genome medicine}, 14(1):68, 2022.

\bibitem{tian2023expanding}
Luyi Tian, Fei Chen, and Evan~Z Macosko.
\newblock The expanding vistas of spatial transcriptomics.
\newblock {\em Nature Biotechnology}, 41(6):773--782, 2023.

\bibitem{sun2024hyperedge}
Xingzhi Sun, Charles Xu, Jo{\~a}o~F Rocha, Chen Liu, Benjamin Hollander-Bodie, Laney Goldman, Marcello DiStasio, Michael Perlmutter, and Smita Krishnaswamy.
\newblock Hyperedge representations with hypergraph wavelets: applications to spatial transcriptomics.
\newblock {\em ArXiv}, pages arXiv--2409, 2024.

\bibitem{Monti2023}
Corrado Monti, Marco Pangallo, Gianmarco De~Francisci~Morales, and Francesco Bonchi.
\newblock On learning agent-based models from data.
\newblock {\em Scientific Reports}, 13(1), June 2023.

\bibitem{mongeon2024spatial}
Blanche Mongeon, Julien H{\'e}bert-Doutreloux, Anudeep Surendran, Elham Karimi, Benoit Fiset, Daniela~F Quail, Logan~A Walsh, Adrianne~L Jenner, and Morgan Craig.
\newblock Spatial computational modelling illuminates the role of the tumour microenvironment for treating glioblastoma with immunotherapies.
\newblock {\em npj Systems Biology and Applications}, 10(1):91, 2024.

\bibitem{chen2018neural}
Ricky~TQ Chen, Yulia Rubanova, Jesse Bettencourt, and David~K Duvenaud.
\newblock Neural ordinary differential equations.
\newblock {\em Advances in neural information processing systems}, 31, 2018.

\bibitem{poli2019graph}
Michael Poli, Stefano Massaroli, Junyoung Park, Atsushi Yamashita, Hajime Asama, and Jinkyoo Park.
\newblock Graph neural ordinary differential equations.
\newblock {\em arXiv preprint arXiv:1911.07532}, 2019.

\bibitem{jin2022multivariate}
Ming Jin, Yu~Zheng, Yuan-Fang Li, Siheng Chen, Bin Yang, and Shirui Pan.
\newblock Multivariate time series forecasting with dynamic graph neural odes.
\newblock {\em IEEE Transactions on Knowledge and Data Engineering}, 35(9):9168--9180, 2022.

\bibitem{bhaskar2024inferring}
Dhananjay Bhaskar, Daniel~Sumner Magruder, Matheo Morales, Edward De~Brouwer, Aarthi Venkat, Frederik Wenkel, Guy Wolf, and Smita Krishnaswamy.
\newblock Inferring dynamic regulatory interaction graphs from time series data with perturbations.
\newblock In {\em Learning on Graphs Conference}, pages 22--1. PMLR, 2024.

\bibitem{vaswani2017attention}
Ashish Vaswani, Noam Shazeer, Niki Parmar, Jakob Uszkoreit, Llion Jones, Aidan~N Gomez, {\L}ukasz Kaiser, and Illia Polosukhin.
\newblock Attention is all you need.
\newblock {\em Advances in neural information processing systems}, 30, 2017.

\bibitem{velivckovic2017graph}
Petar Veli{\v{c}}kovi{\'c}, Guillem Cucurull, Arantxa Casanova, Adriana Romero, Pietro Lio, and Yoshua Bengio.
\newblock Graph attention networks.
\newblock {\em arXiv preprint arXiv:1710.10903}, 2017.

\bibitem{hawrylycz_sea-ad_2024}
Michael Hawrylycz, Eitan~S. Kaplan, Kyle~J. Travaglini, Mariano~I. Gabitto, Jeremy~A. Miller, Lydia Ng, Jennie~L. Close, Rebecca~D. Hodge, Brian Long, Tyler Mollenkopf, Shoaib Mufti, Nicole~M. Gatto, Eric~B. Larson, Paul~K. Crane, Thomas~J. Grabowski, C.~Dirk Keene, and Ed~S. Lein.
\newblock {SEA}-{AD}: {A} multimodal cellular atlas and resource for {Alzheimer}’s disease.
\newblock {\em Nature aging}, 4(10):1331--1334, October 2024.

\bibitem{10xGenomics_Xenium_2025}
{Office of Science and Technology Resources}.
\newblock {10x Genomics Xenium Analyzer}.
\newblock Accessed: Mar. 04, 2025.

\bibitem{liu2024imageflownet}
Chen Liu, Ke~Xu, Liangbo~L Shen, Guillaume Huguet, Zilong Wang, Alexander Tong, Danilo Bzdok, Jay Stewart, Jay~C Wang, Lucian~V Del~Priore, et~al.
\newblock Imageflownet: Forecasting multiscale image-level trajectories of disease progression with irregularly-sampled longitudinal medical images.
\newblock {\em arXiv preprint arXiv:2406.14794}, 2024.

\bibitem{sun2024deep}
Xingzhi Sun, Edward De~Brouwer, Chen Liu, Smita Krishnaswamy, and Ramesh Batra.
\newblock Deep learning unlocks the true potential of organ donation after circulatory death with accurate prediction of time-to-death.
\newblock {\em medRxiv}, pages 2024--11, 2024.

\bibitem{borsari2023chronode}
Beatrice Borsari, Mor Frank, Eve~S Wattenberg, Ke~Xu, Susanna~X Liu, Xuezhu Yu, and Mark Gerstein.
\newblock chronode: A framework to integrate time-series multi-omics data based on ordinary differential equations combined with machine learning.
\newblock {\em bioRxiv}, pages 2023--12, 2023.

\bibitem{villani2008optimal}
C{\'e}dric Villani et~al.
\newblock {\em Optimal transport: old and new}, volume 338.
\newblock Springer, 2008.

\bibitem{santambrogio2015optimal}
Filippo Santambrogio.
\newblock {\em Optimal transport for applied mathematicians}, volume~87.
\newblock Springer, 2015.

\bibitem{tong2020trajectorynet}
Alexander Tong, Jessie Huang, Guy Wolf, David Van~Dijk, and Smita Krishnaswamy.
\newblock Trajectorynet: A dynamic optimal transport network for modeling cellular dynamics.
\newblock In {\em International conference on machine learning}, pages 9526--9536. PMLR, 2020.

\bibitem{matthew2023gillespy2}
S.~Matthew, F.~Carter, J.~Cooper, et~al.
\newblock Gillespy2: A biochemical modeling framework for simulation driven biological discovery.
\newblock {\em Letters in Biomathematics}, 10(1):87--103, 2023.

\end{thebibliography}
\bibliographystyle{unsrt}

\clearpage
\section*{Appendix}

\FloatBarrier 
\begin{algorithm}[H]
\caption{STAGED: Spatio-Temporal Agent-Based Graph Evolution Dynamics}
\label{alg:algorithm}
\footnotesize
\begin{algorithmic}[1]

\Statex \textbf{Inputs:}
    \Statex \hspace{1em} $\bullet$ Spatiotemporal gene expression and location data $\mathcal{D} = \{(x_{c,g,t}, y_{c,t})\}$
    \Statex \hspace{1em} $\bullet$ Prior gene regulatory networks for each cell type: $\mathcal{G}_p = \{G_k\}$
    \Statex \hspace{1em} $\bullet$ Ligand–receptor gene pairs: $\mathcal{P} = \{(g_l, g_r)\}$
    \Statex \hspace{1em} $\bullet$ Cell type annotation function $K(c)$
    \Statex \hspace{1em} $\bullet$ Parameters: GAT weights $W, a$, MLP weights $\theta$
    \Statex \hspace{1em} $\bullet$ Time lags: $\delta_{\text{gg}}, \delta_{\text{gl}}, \delta_{\text{lr}}, \delta_{\text{rg}}$

\Statex

\State \textbf{Step 1: Initialize Cell-Specific Regulatory Graphs}
\ForAll{cell $c$}
    \State Assign cell-type-specific GRN $G_c \gets G_{K(c)}$
    \State Add receptor and ligand nodes to $G_c$ based on known ligand–receptor pairs:
    \ForAll{receptor gene $g_r$}
        \State Add a receptor node $r_{g_r}$ and connect it to $g_r$
        \State Connect $r_{g_r}$ to all other genes (global downstream effect)
    \EndFor
    \ForAll{ligand gene $g_l$}
        \State Add an output ligand node $l_{g_l}$ and connect it to $g_l$
    \EndFor
\EndFor

\Statex

\State \textbf{Step 2: Initialize Model Inputs}
\State Determine warm-up period: $t_{\text{init}} = \max\{\delta_{\text{gg}}, \delta_{\text{gl}}, \delta_{\text{lr}}, \delta_{\text{rg}}\}$
\For{$t = 0$ to $t_{\text{init}}$}
    \State Initialize gene expression values: $\hat{x}_{c,g,t} \gets x_{c,g,t}$
\EndFor

\Statex

\State \textbf{Step 3: Evolve System Over Time}
\For{$t = t_{\text{init}} + 1$ to $T$}
    \ForAll{cell $c$}
        \State Identify spatial neighbors $\mathcal{N}(c)$ based on cell positions at time $t$, initialize the dynamic graph $\tilde{G}_c \leftarrow G_c$
        \ForAll{neighbor cell $c'$ in $\mathcal{N}(c)$}
            \ForAll{ligand–receptor pair $(g_l, g_r)$ in $\mathcal{P}$}
                \State Add ligand node $l_{c'}^{g_l}$ to $\tilde{G}_c$
                \State Connect $l_{c'}^{g_l} \to r_{g_r}$ (intercellular signaling)
            \EndFor
        \EndFor

        \State Assign features to each node in $\tilde{G}_c$ using lagged expression:
            \Statex \hspace{5em} For gene nodes: use $\hat{x}_{c,g,t - \delta_{\text{gg}}}$
            \Statex \hspace{5em} For output ligands: use $\hat{x}_{c,g_l,t - \delta_{\text{gl}}}$
            \Statex \hspace{5em} For receptor nodes: use $\hat{x}_{c,g_r,t - \delta_{\text{rg}}}$
            \Statex \hspace{5em} For input ligands from neighbors: use $\hat{x}_{c',g_l,t - \delta_{\text{lr}}}$

        \State Apply GAT layer on $\tilde{G}_c$ to get attention weights and embeddings
        \State Use MLP to update hidden state of each gene
        \State Integrate dynamics using a neural ODE to predict next timepoint
    \EndFor
\EndFor

\Statex

\State \textbf{Return:} Predicted gene expression over time: $\{\hat{x}_{c,g,t}\}$ and learned attention weights $\alpha$
\end{algorithmic}
\end{algorithm}

\end{document}